\documentclass[sigconf]{acmart}

\renewcommand\footnotetextcopyrightpermission[1]{} 
\usepackage{pdfpages}
\usepackage{framed}

\begin{document}

\title{FA Team at the NTCIR-17 UFO Task}

\author{Yuki Okumura}
\affiliation{
    \institution{Fast Accounting Co., Ltd.}
    \country{Japan}
}
\email{okumura.yuki@fastaccounting.co.jp}
\author{Masato Fujitake}
\affiliation{
    \institution{FA Research, Fast Accounting Co., Ltd.}
    \country{Japan}
}
\email{fujitake@fastaccounting.co.jp}

\begin{abstract}
The FA team participated in the Table Data Extraction (TDE) and Text-to-Table Relationship Extraction (TTRE) tasks of the NTCIR-17 Understanding of Non-Financial Objects in Financial Reports (UFO). 
This paper reports our approach to solving the problems and discusses the official results.
We successfully utilized various enhancement techniques based on the ELECTRA language model to extract valuable data from tables. 
Our efforts resulted in an impressive TDE accuracy rate of 93.43\%, positioning us in second place on the Leaderboard rankings. 
This outstanding achievement is a testament to our proposed approach's effectiveness. 
In the TTRE task, we proposed the rule-based method to extract meaningful relationships between the text and tables task and confirmed the performance.
\end{abstract}


\maketitle
\pagestyle{plain} 

\section*{Team Name}
FA

\section*{Subtasks}
Table Data Extraction subtask (Japanese) \\
Text-to-Table Relationship Extraction  subtask (Japanese)

\section{Introduction}  \label{sec:intro}

In natural language processing and information retrieval, extracting valuable data from tables and recognizing the connection between the written content and table representations is crucial.
These tasks are fundamental in several domains, including data analysis, decision-making, and knowledge extraction, particularly in specific fields like financial reporting.
As financial reporting becomes increasingly inundated with non-financial information, it is imperative for stakeholders, investors, and financial analysts to effectively understand and interpret the data for analysis.
During the 17th NTCIR Conference on Understanding Non-financial Objects in Financial Reports (UFO)\cite{ntcir17-ufo-overview}, two tasks were used to encourage the development of practical algorithms for extracting and interpreting table-related data. 
These tasks, known as Table Data Extraction (TDE) and Text-to-Table Relational Extraction (TTRE), provide a structured framework for studying and addressing the complexities involved in these processes.

We share our practical approach to conducting these tasks in this paper. 
For the TDE task, we leveraged advanced techniques using the powerful language model called ELECTRA~\cite{clark2020ELECTRA}.
Furthermore, we introduced a post-correction method based on the Levenshtein distance for the language model's output to reduce errors.
This approach resulted in an impressive accuracy rate of 93.43\%, showcasing its effectiveness.
For the TTRE task, we proposed a rule-based method to address the issue and validated the method.

The structure of this paper is as follows: Section~\ref{sec:relatedwork} covers related work, Section~\ref{sec:method} gives a detailed overview of our method, Section~\ref{sec:experiments} presents results and analysis, and Section~\ref{sec:conclusion} concludes with a discussion of the implications of our work and future research directions.
\section{Related Work}  \label{sec:relatedwork}

\noindent
\textbf{Table Data Extraction (TDE).} 
Extracting information from tabular data is complex and challenging because it requires considering the information in the cells and the surrounding information.
Therefore, a method that treats data not as individual cells but as a group of rows has been proposed to handle table structures.
In previous research, cells are connected using special symbols, and treated as a single textual information, considered an input method to the language model~\cite{chen2019TabFactA}.
Treating table data as sequence data makes it possible to classify information in a row effectively. 
Recently, a method involves approaching cell information as a Named Entity Recognition task~\cite{souza2020portuguese}. 
The words can be accurately classified by treating the table data as sequence data and adapting NER as sequence labeling.

We have developed a novel approach to analyzing table data based on previous research findings. 
We treat tables as sequence data and extract information accordingly. 
Unlike existing studies,  our method differs in two key respects. 
First, we combine both data of the classification-target cell and the corresponding entire row.
It allows us to consider the broader context of the whole row, thereby enhancing the accuracy of our analysis.
Second, we treated the task as a cell-by-cell classification rather than an NER task to focus on a target cell.
We have conducted a comparison experiment between the proposed method and the NER approach to demonstrate the effectiveness of the proposed method.

To handle the sequence data, we utilized a language model that models language using word occurrence probabilities. 
Recent years have seen the emergence of various methods, including BERT~\cite{devlin2018bert}. 
We employed ELECTRA~\cite{clark2020ELECTRA}, an extension of BERT~\cite{devlin2018bert} to extract table information. 
While BERT performs pre-training by randomly filling in sentences, ELECTRA~\cite{clark2020ELECTRA} proposes a more sophisticated way of filling in sentences by focusing on each word and guessing which parts have been filled in by language models, thus improving accuracy compared to BERT. 
For table information extraction, we adapted ELECTRA, which better handles the meaning of each word than the whole sentence, as the target of table information is a short string of a few words.
In a preliminary experiment, we conducted a verification of ELECTRA's effectiveness as compared to BERT.

\noindent
\textbf{Text-to-Table Relationship Extraction (TTRE).} 
To link sentences and tabular data, one can consider it a specialized case of the Entity Linking task. 
The task aims to connect a given text with a representation of knowledge. 
Two primary approaches, rule-based and machine learning-based methods, have been proposed for this task.
Rule-based methods have been proposed to link elements based on similarity measured by Levenshtein Distance~\cite{ohsawa2014PopularityPF, levenshtein1966binary}.
The Levenshtein distance measures the similarity of two strings by counting edit operations to make them identical.
However, rule-based methods have issues, such as a lower matching rate due to differences in notation. 
Therefore, machine learning-based methods have recently been used to improve robustness~\cite{mehwish2022entitylinkingsurvey}.
In this study, since this task is a particular case dealing with Entity Linking table data, we adopted a rule-based approach that eliminates complexity to clarify future issues.

\section{Methods}  \label{sec:method}
\subsection{Table Data Extraction}

The task is to understand the structure of the tables in the financial report.
More specifically, it is to predict the role of each cell in the table.
The input is a report written in HTML, and the output is an ID assigned to every table cell in the report and its categories (Metadata, Header, Attribute, Data) for the cell.
Structuring the tables can be used for numerical comparisons between firms.

To address this issue, we use a language model-based text classification model to predict the category of each table cell.
Instead of outputting raw model results, we utilize a post-correction technique to improve accuracy.
Details are described below.

\begin{figure}[htp]
	\centering
	\includegraphics[width=1.00\columnwidth, keepaspectratio]{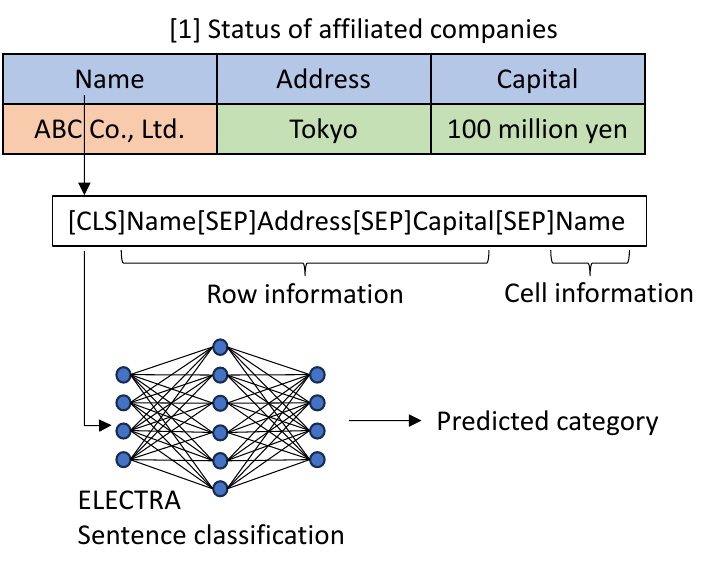}
	\caption{
            The architecture of the proposed method.
            Sentence classification is performed by inputting a cell and the line in which the cell is located.}
	\label{fig:overall_model_lm}
\end{figure}

\subsubsection{Cell Classification Method}
We utilized contextual information surrounding the target cells of the table to improve cell classification. 
Our classification pipeline, depicted in Figure~\ref{fig:overall_model_lm}, incorporates a language model. 
Due to the brevity of cell text, predicting meaning based solely on text is challenging. 
To address this issue and account for table structure, we treated an entire row and corresponding cells as a single text for classification.
Each cell's information is combined using a specific marker to indicate the joining of cells and the classification of cell information. 
A special token \texttt{[SEP]} is added to each cell to express the division of cells in the row representation, as shown in Figure~\ref{fig:overall_model_lm}. 
However, since the maximum input token length for the language model is $L$, if the input target is longer than $L$ tokens, the sentence is truncated, and only $L$ tokens are used.

\begin{figure}[htp]
	\centering
	\includegraphics[width=1.00\columnwidth, keepaspectratio]{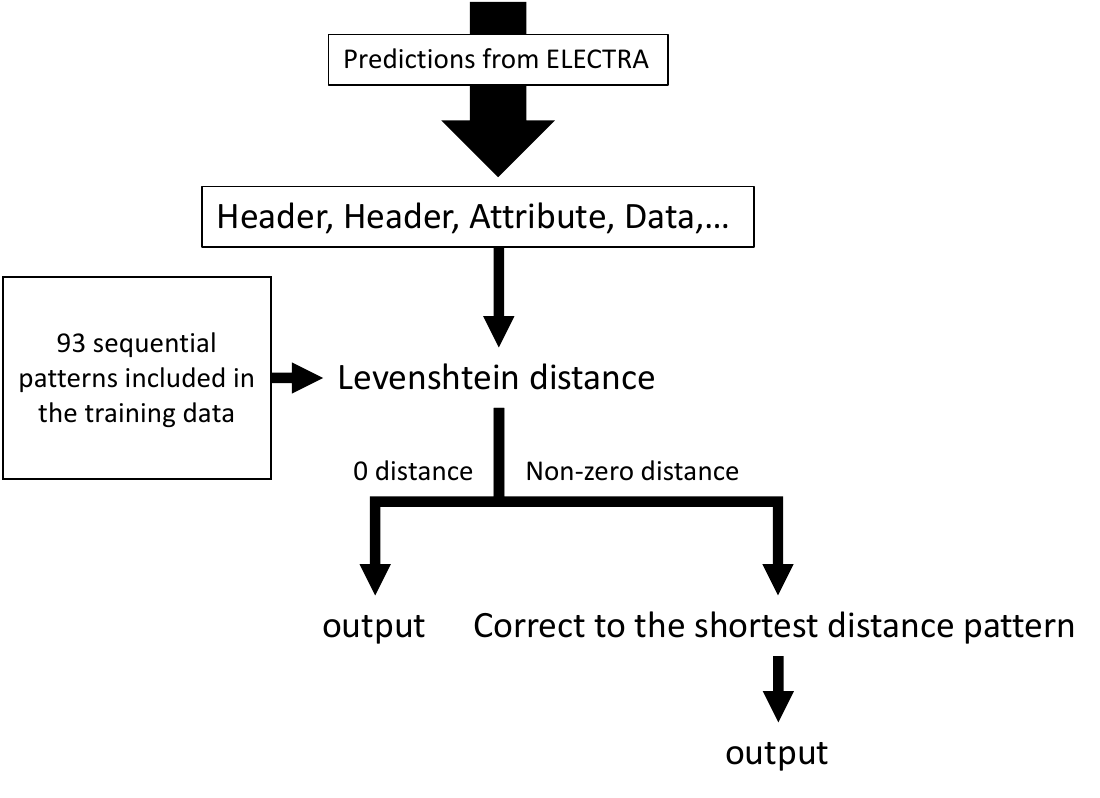}
	\caption{
The pipeline of the proposed correction method.
After classifying all cells in a row, the current pattern is matched against any existing ones, and the most similar one is selected.
	}
	\label{fig:overall_model_correction}
     \vspace*{-1.00\baselineskip}
\end{figure}

\subsubsection{Post-Correction Method}
We propose a post-correction method to improve the classification accuracy of the language model's results. 
The output results classified cell by cell may be inconsistent across an entire row.
Therefore, we propose a post-correction method based on edit distance to ensure consistency.
The pipeline is described in Figure~\ref{fig:overall_model_correction}.

The cell categories predicted by the language model are collected row by row to check whether the sequential pattern was present in the training data.
As shown in Figure~\ref{fig:overall_model_correction}, pattern matching is performed using the Levenshtein distance.
The 93 patterns were extracted from all the DryRun and Formal Run train dataset row series distributed in the UFO task.
Suppose the output sequence does not match the patterns. 
In that case, it is corrected to one that can be matched at minimum cost by a Levenshtein distance editing operation.
However, if the length of a predicted sequence is longer than the patterns, the add and delete operations are ignored, and only replacements are performed.

\subsection{Text-to-Table Relationship Extraction}
The objective is to match tables and their corresponding explanations in financial reports.
The task involves analyzing an HTML-formatted report and generating related sets of explanatory text and table elements (such as headers, data, and schemas).
Extracting supplemental information is necessary for tables to effectively describe structured information.
This enables a comprehensive analysis of the enterprise.

We present a rule-based approach to linking complex tables with textual representations. 
Understanding both text and table information poses a challenge. 
We evaluate the effectiveness of a rule-based approach as a baseline without a machine learning method. 
The following pipeline is used to output results from the input.

\begin{enumerate}
    \item Extract the table from the HTML and separate the text from the table. 
        We remark that it is necessary to maintain and extract that information because rowspan and colspan in the HTML can mess up the table's structure.
        In extracting the tables, we recorded the order in which the tables and descriptions appeared and stored them as structural data.
        We use this recorded data to extract the table that precedes the description when we select the table in the following steps.

    \item Preprocess to increase the match rate.
        Specifically, the sentences to be extracted are divided based on particles and parentheses.
    \item Identify the Name.
        In the table structure, the top two rows and the left two columns are often not numerical values but rather headers or other keys.
        Therefore, cells in this area are candidates for assigning Names. 
        The percentage of matches with the segmented sentences is measured for each cell.
        The cell is assigned as Name if the match rate is greater than 70\% in the Levenshtein distance.
    \item Identify Value. 
        There are two ways to identify a value.
        The usual pattern is that if two Names are found in a candidate area, the cell at the intersection of the row and column of the found cell is identified as Value.
        Exceptionally, only one Name is detected in a candidate area. 
        In that case, all row or column elements corresponding to the detected cell are retained as candidates for Value.
    \item Narrow down the candidates of Value.
        Since Value is basically expressed numerically, the candidates should be numerical values.
        As a candidate judgment, at least 50\% of the text of the value item must contain numeric characters. 
        Cells not meeting this criterion are removed from the value candidates and labeled as ``etc.'', defined as a misc category.
\end{enumerate}

\section{Experiments} \label{sec:experiments}
\subsection{Table Data Extraction}
\subsubsection{Implementation Details}

\begin{table}[htbp]
\caption{Category distribution in TDE dataset.}
\label{table:tde_dataset}
\resizebox{1.0\columnwidth}{!}{%
\begin{tabular}{c|r|c|c|c|c}
\toprule
Dataset    & Total & header                     & attribute                  & data                       & metadata                 \\ \midrule
train  & 78,926                      & 16,568 & 14430 & 47,680 & 248 \\ 
test   & 45,499                      & -                          & -                          & -                          & -                        \\ \bottomrule
\end{tabular}
}
\end{table}

Table~\ref{table:tde_dataset} provides detailed information about the dataset. 
Unfortunately, we could not confirm the number of categories in the test data, as it was not disclosed.
The evaluation results are presented as the macro average of accuracy per table.
The Japanese-language model ``izumi-lab/electra-base-japanese-discriminator''\footnote{\url{https://huggingface.co/izumi-lab/electra-base-japanese-discriminator}} was used as the pre-trained ELECTRA model.
The max token length $L$ is set to 128.
Fine-tuning was performed using the following parameters, and the model with the highest accuracy was adopted through 5-part cross-validation.
The optimizer was Adam, and training was performed with a learning rate of 1e-5.
The training period was five epochs.

\begin{table}[htbp]
\caption{Performance comparison with the state-of-the-art
methods on TDE.
}
\label{table:tde_overall_result}
\resizebox{0.40\columnwidth}{!}{%
\begin{tabular}{c|c}
\toprule
Team                                                                                          & F1-score \\ \midrule
KSU & \textbf{0.9537} \\
FA (Ours) & 0.9343 \\
OUC & 	0.9217 \\
jpxiteam & 0.8287 \\
TO & 0.7981 \\ \bottomrule

\end{tabular}
}
    \vspace*{-1.00\baselineskip}
\end{table}

\subsubsection{Main Result}
Table~\ref{table:tde_overall_result} displays team names and accuracy on the Formal Run Leaders Board\footnote{\url{https://sites.google.com/view/ntcir17-ufo/leaderboard?authuser=0}}. 
Our method achieved 93.43\%, making us second with competitive accuracy. 
KSU employed a method considering the table structure based on BERT, while OUC uses BERT Large~\cite{devlin2018bert}.
The jpxiteam uses ChatGPT, and the TO uses rule-based methods. 
The proposed method achieves high accuracy using the Electra-based language model with fine-tuning and post-processing.

\begin{table*}[htbp]
\caption{Ablation studies on TDE.
}
\label{table:tde_result_analysis}
\resizebox{1.5\columnwidth}{!}{%
\begin{tabular}{c|ccc|c}
\toprule
Approach                & Target Cell Information & Row Information & Post Correction Method & F1-score \\ \midrule
Sentence Classification & \checkmark       & \checkmark  & \checkmark      & \textbf{0.9343}   \\ 
Sentence Classification & \checkmark       & \checkmark  &                 & 0.9321   \\ 
Sentence Classification & \checkmark       &             & \checkmark      & 0.9236   \\ 
Sentence Classification & \checkmark       &             &                 & 0.9140   \\ 
\hline
Sequence Labeling (NER) &                  & \checkmark  &                 & 0.7937   \\  \bottomrule
\end{tabular}
}
\end{table*}

\begin{figure}[htp]
	\centering
 	\includegraphics[width=1.00\columnwidth, keepaspectratio]{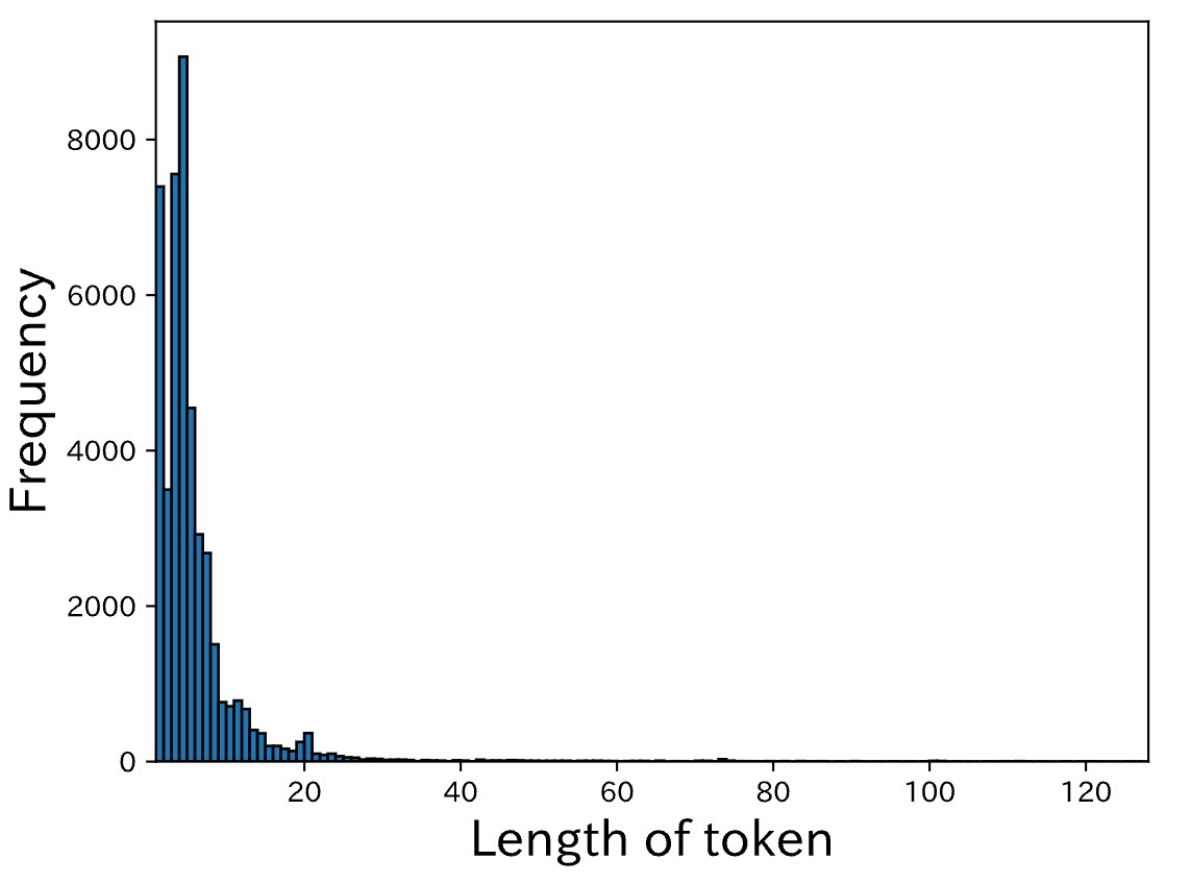}
	\caption{
Distribution of token lengths in the cells of the table.
Most of the content within the cells contains only a limited number of tokens, indicating that the text is brief and comprises individual words rather than entire sentences.
	}
    \label{fig:cell_length_histgram}
        \vspace*{-1.00\baselineskip}
\end{figure}

\subsubsection{Detailed Analysis}
We conducted a detailed analysis to see what effectively extracts the table information.
Here, we analyzed each cell and the component analysis of the method.

First, we analyzed the token length of each cell handled by the language model because the length of the tokens affects the semantic information that the language model obtains from the text.
The histograms of the length and frequency of the tokens in each cell are shown in Figure~\ref{fig:cell_length_histgram}.
It also shows that the number of tokens in each cell is concentrated into a few tokens.
This suggests the language model may not obtain enough text to ensure context when dealing with cells.

Next, we conducted an ablation study to analyze what affects performance in the TDE task.
In this experiment, we validated the NER approach of sequence modeling with ELECTRA+CRF to confirm the effectiveness of the cell-by-cell text classification approach.
In NER, each line was used as an input, and as in text classification, cell delimiters were represented by special tokens \texttt{[SEP]}, the category of each token between \texttt{[SEP]} was predicted, and if multiple tokens in a cell made separate predictions, the later prediction was given priority.
The experiment results are shown in Table~\ref{table:tde_result_analysis}.

We confirm that the classification approach with a single cell achieved 91\% accuracy compared to 79\% accuracy with NER and that the extraction of table information as a classification task is 12 points more accurate.
In addition, based on the analysis of token length in cells, it was confirmed that even text with a few tokens can achieve a certain degree of accuracy in language models. 
At first, it was thought that incorporating lengthy sequences of tokens and contextual information, such as complete sentences, would increase accuracy. 
It was proven that the assumption may only sometimes be accurate.
When comparing text classification that considers cells alone and cells plus entire rows, the accuracy is improved by two points to 93\%.
Therefore, ensuring context in the TDE task is essential, as a few tokens alone cannot consider the surrounding information.
To enhance output consistency, post-processing was implemented. We verified the efficacy of post-processing, resulting in improved accuracy.

For future accuracy improvements, we believe it is crucial to use domain-specific pre-trained models~\cite{choi2022domain}.
We used an existing pre-trained language model to implement our proposed method, a model trained by a generic corpus such as Wikipedia.
However, in this case, the text sentences in the tables are shorter than general sentences. 
Moreover, in some cases, more than the generic content is required.
Therefore, using a pre-trained model specific to table representation, rather than a pre-trained model trained on generic data and for generic purposes, will allow for more accurate classification.

\subsection{Text-to-Table Relationship Extraction}
\subsubsection{Implementation Details}
The data used for the test is a total of 25 HTML files, and we answer cells corresponding to a total of 11,867 cell descriptions.
The accuracy in evaluation is measured by the F-1 score.
The three perspectives are considered:  ``Name'', ``Value'', and their average.

\begin{table}[htbp]
\caption{Text to Table Relationship Extraction result.}
\label{table:ttre_result}
\resizebox{0.7\columnwidth}{!}{%
\begin{tabular}{c|cc|c}
\toprule
Method & Name   & Value  & Total  \\ \midrule
Rule-based (Ours) & 0.0341 & 0.0131 & 0.0236 \\
Random            & 0.0008 & 0.0004 & 0.0006 \\
\bottomrule
\end{tabular}
}
    \vspace*{-1.00\baselineskip}
\end{table}

\subsubsection{Main Result and Discussion}
The experiment results are shown in table~\ref{table:ttre_result}.
Also shown are the results of the random prediction output using the code provided by the organizer of the UFO task.
We see that our result of an overall F1 score is 2.4\%.
Also, Name and Value scored 3.4\% and 1.3\%, respectively, confirming that Name achieves better on a rule basis.

From the experiments, the rule-based approach can extract relationships in several patterns, which is verified by the comparison with random prediction. 
Still, the overall accuracy could be better and needs further improvement for practical usage.
First, the Levenshtein distance was used to determine the match rate in the rule-based approach. 
Currently, we use a fixed match rate for all cases. 
However, changing the match rate based on attributes adaptively, such as Name and Value, is necessary.
In addition, we regarded the text as a numerical expression if 50\% of the text was a numerical expression, etc., otherwise. 
Thus, the classification accuracy needed to be higher.
For instance, a binary classification model should be introduced to separate the data and etc elements.
Moreover, it is necessary to consider introducing a machine-learning approach.
Specifically, although we extracted the data when there is a match in terms of ``Name'', we believe it may be practical to increase the accuracy by using a method that measures the similarity of feature values.

\section{Conclusions} \label{sec:conclusion}

We presented our methodology and the results of our participation in the task of understanding non-financial objects (UFO) in NTCIR-17 financial reports, specifically in table data extraction (TDE) and text-to-table relationship extraction (TTRE).
The fact that our language model and post-processing method for table data extraction achieved an accuracy of 93.43\% and ranked second on the Leaderboard demonstrates the success and robustness of our proposed approach.
In addition, the detailed analysis demonstrated the effectiveness of our approach by comparing it with other methods, such as named entity recognition approaches.
For text-to-table relationship extraction, our rule-based approach has established a baseline for future research.
With these results, we discussed future research directions, including the need for domain-specific pre-trained models and the introduction of machine-learning approaches.


\bibliographystyle{ACM-Reference-Format}
\bibliography{article}


\begin{thebibliography}{9}


\ifx \showCODEN    \undefined \def \showCODEN     #1{\unskip}     \fi
\ifx \showDOI      \undefined \def \showDOI       #1{#1}\fi
\ifx \showISBNx    \undefined \def \showISBNx     #1{\unskip}     \fi
\ifx \showISBNxiii \undefined \def \showISBNxiii  #1{\unskip}     \fi
\ifx \showISSN     \undefined \def \showISSN      #1{\unskip}     \fi
\ifx \showLCCN     \undefined \def \showLCCN      #1{\unskip}     \fi
\ifx \shownote     \undefined \def \shownote      #1{#1}          \fi
\ifx \showarticletitle \undefined \def \showarticletitle #1{#1}   \fi
\ifx \showURL      \undefined \def \showURL       {\relax}        \fi
\providecommand\bibfield[2]{#2}
\providecommand\bibinfo[2]{#2}
\providecommand\natexlab[1]{#1}
\providecommand\showeprint[2][]{arXiv:#2}

\bibitem[Chen et~al\mbox{.}(2020)]%
        {chen2019TabFactA}
\bibfield{author}{\bibinfo{person}{Wenhu Chen}, \bibinfo{person}{Hongmin Wang},
  \bibinfo{person}{Jianshu Chen}, \bibinfo{person}{Yunkai Zhang},
  \bibinfo{person}{Hong Wang}, \bibinfo{person}{Shiyang Li},
  \bibinfo{person}{Xiyou Zhou}, {and} \bibinfo{person}{William~Yang Wang}.}
  \bibinfo{year}{2020}\natexlab{}.
\newblock \showarticletitle{Tabfact: A large-scale dataset for table-based fact
  verification}. In \bibinfo{booktitle}{\emph{Proceedings of the International
  Conference on Learning Representations}}.
\newblock


\bibitem[Choi et~al\mbox{.}(2022)]%
        {choi2022domain}
\bibfield{author}{\bibinfo{person}{Dongha Choi}, \bibinfo{person}{HongSeok
  Choi}, {and} \bibinfo{person}{Hyunju Lee}.} \bibinfo{year}{2022}\natexlab{}.
\newblock \showarticletitle{Domain Knowledge Transferring for Pre-trained
  Language Model via Calibrated Activation Boundary Distillation}. In
  \bibinfo{booktitle}{\emph{Proceedings of the Annual Meeting of the
  Association for Computational Linguistics}}. \bibinfo{pages}{1658--1669}.
\newblock


\bibitem[Clark et~al\mbox{.}(2020)]%
        {clark2020ELECTRA}
\bibfield{author}{\bibinfo{person}{Kevin Clark}, \bibinfo{person}{Minh-Thang
  Luong}, \bibinfo{person}{Quoc~V. Le}, {and} \bibinfo{person}{Christopher~D.
  Manning}.} \bibinfo{year}{2020}\natexlab{}.
\newblock \showarticletitle{ELECTRA: Pre-training Text Encoders as
  Discriminators Rather Than Generators}. In
  \bibinfo{booktitle}{\emph{Proceedings of the International Conference on
  Learning Representations}}.
\newblock


\bibitem[Devlin et~al\mbox{.}(2018)]%
        {devlin2018bert}
\bibfield{author}{\bibinfo{person}{Jacob Devlin}, \bibinfo{person}{Ming-Wei
  Chang}, \bibinfo{person}{Kenton Lee}, {and} \bibinfo{person}{Kristina
  Toutanova}.} \bibinfo{year}{2018}\natexlab{}.
\newblock \showarticletitle{Bert: Pre-training of deep bidirectional
  transformers for language understanding}. In
  \bibinfo{booktitle}{\emph{Proceedings of the North American Chapter of the
  Association for Computational Linguistics}}.
\newblock


\bibitem[Kimura et~al\mbox{.}(2023)]%
        {ntcir17-ufo-overview}
\bibfield{author}{\bibinfo{person}{Yasutomo Kimura}, \bibinfo{person}{Hokuto
  Ototake}, \bibinfo{person}{Kazuma Kadowaki}, \bibinfo{person}{Takahito
  Kondo}, {and} \bibinfo{person}{Makoto~P. Kato}.}
  \bibinfo{year}{2023}\natexlab{}.
\newblock \showarticletitle{Overview of the NTCIR-17 UFO Task}.
\newblock \bibinfo{journal}{\emph{Proceedings of The 17th NTCIR Conference}}
  (\bibinfo{date}{12} \bibinfo{year}{2023}).
\newblock


\bibitem[Levenshtein et~al\mbox{.}(1966)]%
        {levenshtein1966binary}
\bibfield{author}{\bibinfo{person}{Vladimir~I Levenshtein} {et~al\mbox{.}}}
  \bibinfo{year}{1966}\natexlab{}.
\newblock \showarticletitle{Binary codes capable of correcting deletions,
  insertions, and reversals}. In \bibinfo{booktitle}{\emph{Soviet physics
  doklady}}, Vol.~\bibinfo{volume}{10}. \bibinfo{pages}{707--710}.
\newblock


\bibitem[Ohsawa and Matsuo(2014)]%
        {ohsawa2014PopularityPF}
\bibfield{author}{\bibinfo{person}{Shohei Ohsawa} {and} \bibinfo{person}{Yutaka
  Matsuo}.} \bibinfo{year}{2014}\natexlab{}.
\newblock \showarticletitle{Popularity Prediction for Entities on SNS Using
  Semantic Relations}.
\newblock \bibinfo{journal}{\emph{Transactions of The Japanese Society for
  Artificial Intelligence}}  \bibinfo{volume}{29} (\bibinfo{year}{2014}),
  \bibinfo{pages}{469--482}.
\newblock


\bibitem[Sevgili et~al\mbox{.}(2022)]%
        {mehwish2022entitylinkingsurvey}
\bibfield{author}{\bibinfo{person}{{\"O}zge Sevgili}, \bibinfo{person}{Artem
  Shelmanov}, \bibinfo{person}{Mikhail Arkhipov}, \bibinfo{person}{Alexander
  Panchenko}, {and} \bibinfo{person}{Chris Biemann}.}
  \bibinfo{year}{2022}\natexlab{}.
\newblock \showarticletitle{Neural entity linking: A survey of models based on
  deep learning}.
\newblock \bibinfo{journal}{\emph{Semantic Web}} \bibinfo{volume}{13},
  \bibinfo{number}{3} (\bibinfo{year}{2022}), \bibinfo{pages}{527--570}.
\newblock


\bibitem[Souza et~al\mbox{.}(2020)]%
        {souza2020portuguese}
\bibfield{author}{\bibinfo{person}{Fábio Souza}, \bibinfo{person}{Rodrigo
  Nogueira}, {and} \bibinfo{person}{Roberto Lotufo}.}
  \bibinfo{year}{2020}\natexlab{}.
\newblock \showarticletitle{Portuguese Named Entity Recognition using
  BERT-CRF}.
\newblock \bibinfo{journal}{\emph{arXiv preprint arXiv:1909.10649}}.
\newblock


\end{thebibliography}

\end{document}